\documentclass[letterpaper]{article}
\usepackage{aaai25}
\usepackage{times}
\usepackage{helvet}
\usepackage{courier}
\usepackage[hyphens]{url}
\usepackage{graphicx}
\usepackage[numbers,sort&compress]{natbib}
\usepackage{caption}
\usepackage{amsmath,amssymb}
\usepackage{booktabs}
\usepackage{algorithm}
\usepackage{algorithmic}
\nocopyright  
\title{Crisis-Resilient Portfolio Management via Graph-based Spatio-Temporal Learning}

\author{
    Zan Li,
    Rui Fan
}

\affiliations{
    Rensselaer Polytechnic Institute\\USA
}

\begin{document}
\maketitle

\begin{abstract}
Financial time series forecasting faces a fundamental challenge: predicting optimal asset allocations requires understanding regime-dependent correlation structures that transform during crisis periods. Existing graph-based spatio-temporal learning approaches rely on predetermined graph topologies—correlation thresholds, sector classifications—that fail to adapt when market dynamics shift across different crisis mechanisms: credit contagion, pandemic shocks, or inflation-driven selloffs.

We present CRISP (Crisis-Resilient Investment through Spatio-temporal Patterns), a graph-based spatio-temporal learning framework that encodes spatial relationships via Graph Convolutional Networks and temporal dynamics via BiLSTM with self-attention, then learns sparse structures through multi-head Graph Attention Networks. Unlike fixed-topology methods, CRISP discovers which asset relationships matter through attention mechanisms, filtering 92.5\% of connections as noise while preserving crisis-relevant dependencies for accurate regime-specific predictions.

Trained on 2005--2021 data encompassing credit and pandemic crises, CRISP demonstrates robust generalization to 2022--2024 inflation-driven markets—a fundamentally different regime—by accurately forecasting regime-appropriate correlation structures. This enables adaptive portfolio allocation that maintains profitability during downturns, achieving Sharpe ratio 3.76: 707\% improvement over equal-weight baselines and 94\% improvement over static graph methods. Learned attention weights provide interpretable regime detection, with defensive cluster attention strengthening 49\% during crises versus 31\% market-wide—emergent behavior from learning to forecast rather than imposing assumptions.
\end{abstract}

\section{Introduction}

Financial markets exhibit regime-dependent behavior where correlation structures transform unpredictably during crises. \textbf{The core challenge:} maintaining profitable portfolios requires \textit{forecasting} which asset relationships will strengthen or weaken as market regimes shift—a fundamentally different problem from modeling relationships that already exist. The 2008 financial crisis propagated through credit networks via counterparty risk~\cite{elliott2014financial}, COVID-19 through consumer demand shocks~\cite{baker2020covid}, and 2022 inflation through monetary tightening affecting rate-sensitive assets~\cite{kirilenko2017flash}. Each crisis mechanism created distinct correlation patterns requiring different relationship structures for accurate prediction, yet existing methods impose predetermined topologies that fail when new crisis types emerge.

\textbf{Limitations of existing approaches.} Classical portfolio optimization from Markowitz~\cite{markowitz1952portfolio} to risk parity~\cite{qian2005risk} assumes stable correlations, breaking catastrophically when dependencies spike during crises~\cite{cont2001empirical,hamilton1989new}. Recent deep learning advances treat the portfolio problem as pure time series forecasting: LSTM networks capture temporal dependencies~\cite{zhou2021informer}, Temporal Fusion Transformers~\cite{lim2021temporal} enable multi-horizon prediction with interpretable attention, and architectures like N-BEATS~\cite{oreshkin2020nbeats} achieve strong univariate performance. However, these methods model assets independently, missing the network contagion dynamics that define crisis propagation~\cite{diebold2014network}—when one asset crashes, which others follow?

Graph neural networks address this by modeling relationships explicitly. GCNs~\cite{kipf2017gcn} and GATs~\cite{velickovic2018gat} have shown promise in graph-based spatio-temporal learning for traffic (DCRNN) and multivariate time series. Financial applications leverage correlation thresholds~\cite{mantegna1999hierarchical}, sector classifications~\cite{zhang2020stock}, knowledge graphs~\cite{feng2019temporal}, or hypergraphs~\cite{sawhney2021stock,li2023tra}. \textbf{Critical limitation:} all these methods construct graph topology through predetermined rules before training begins. When 2022 inflation caused financial-technology correlations to surge from 0.35 to 0.67, static correlation thresholds (e.g., 0.5) missed this regime shift entirely. A connection with correlation 0.45—below the threshold—might indicate crucial hedging relationships during crises but gets filtered as noise by fixed rules. \textit{The fundamental flaw: deciding which relationships matter before observing how different regimes unfold prevents adaptation to novel crisis mechanisms.}

\textbf{Our approach.} Rather than imposing graph structure through correlation thresholds or sector rules, we reframe the problem: \textit{can we learn which relationships to forecast as important in each regime, purely from optimizing portfolio performance?} We initialize with all possible connections (fully connected graph: 156 edges for 13 stocks) and let the model discover through differentiable attention which connections improve forecasting accuracy. During training on 2005--2021 data encompassing credit contagion (2008) and pandemic demand shocks (2020), the model learns to assign high attention ($>$0.3) to crisis-relevant relationships and low attention ($<$0.1) to noise, naturally producing sparse structures (92.5\% filtering, $\sim$12 preserved edges) without imposing thresholds. When tested on 2022--2024 inflation-driven markets—a fundamentally different crisis mechanism never seen during training—this learned graph discovery enables accurate regime-specific forecasting: Sharpe ratio 3.76 versus 1.94 for static graph methods (94\% improvement), isolating the value of learned versus predetermined topologies for cross-regime generalization.

\subsection{Our Contributions}

We present CRISP (Crisis-Resilient Investment through Spatio-temporal Patterns), advancing regime-aware forecasting through three key innovations:

\begin{itemize}
\item \textbf{Learnable graph sparsity for structure forecasting.} We discover that portfolio-relevant market structure is far sparser than correlation matrices suggest. By encoding spatio-temporal patterns through Graph Convolutional Networks (spatial relationships) and BiLSTM with self-attention (temporal dynamics), then learning through multi-head Graph Attention Networks which connections improve forecasting accuracy, CRISP naturally filters 92.5\% of edges from fully connected graphs (156 edges) while preserving $\sim$12 meaningful relationships. Unlike threshold-based methods that discard connections based on historical correlation magnitude, attention-based filtering is differentiable and adapts based on \textit{forecasting performance}—learning which relationships matter for prediction in each regime. This paradigm shift from imposing to discovering structure proves essential for cross-regime generalization.

\item \textbf{Cross-regime forecasting through attention-based adaptation.} Training on mechanistically distinct crises (2008 credit contagion, 2020 pandemic demand shocks) enables generalization to novel mechanisms (2022 inflation monetary tightening) through accurate regime-specific structure prediction. CRISP achieves Sharpe 3.76 on 2022--2024: 707\% over equal-weight baselines and 94\% over static graph methods trained on identical data. This 94\% gap isolates the value of learned versus predetermined topologies—static graphs optimized for historical patterns fail when crisis mechanisms change; our attention-based discovery adapts by forecasting which relationships will become important under regime uncertainty.

\item \textbf{Interpretable regime detection through attention evolution.} Attention weights provide transparent real-time forecasting of relationship importance. During the 2022 crisis, defensive cluster attention (8 protective stocks: WMT, CL, JNJ, KR, AWK, ABT, NU, XEL) rose from 25\% pre-crisis average to 38\% peak (+49\% relative increase) while market average increased only 13\% to 17\% (+31\%). This selective strengthening—defensive +49\% versus market +31\%—emerges automatically from learning to optimize portfolio performance, without explicit supervision on crisis periods or defensive stock labels. Post-crisis, attention returns to baseline, demonstrating reversible adaptation: the model correctly forecasts both crisis concentration and recovery diversification. Unlike post-hoc explainability methods~\cite{lundberg2017shap}, attention weights are intrinsic to the forward pass, enabling practitioners to examine which relationships the model forecasts as important at each decision point.
\end{itemize}

Our learnable graph paradigm extends beyond portfolio management to any domain with latent, evolving network structures: IoT sensor networks where device relationships change with operational conditions, traffic systems where road usage patterns shift across peak/off-peak regimes, climate modeling where atmospheric connections vary seasonally, epidemic forecasting where contact patterns evolve during outbreaks—domains where relationships must be discovered and forecasted rather than assumed static.

\section{Methodology}

\begin{figure}[t]
\centering
\includegraphics[width=\linewidth]{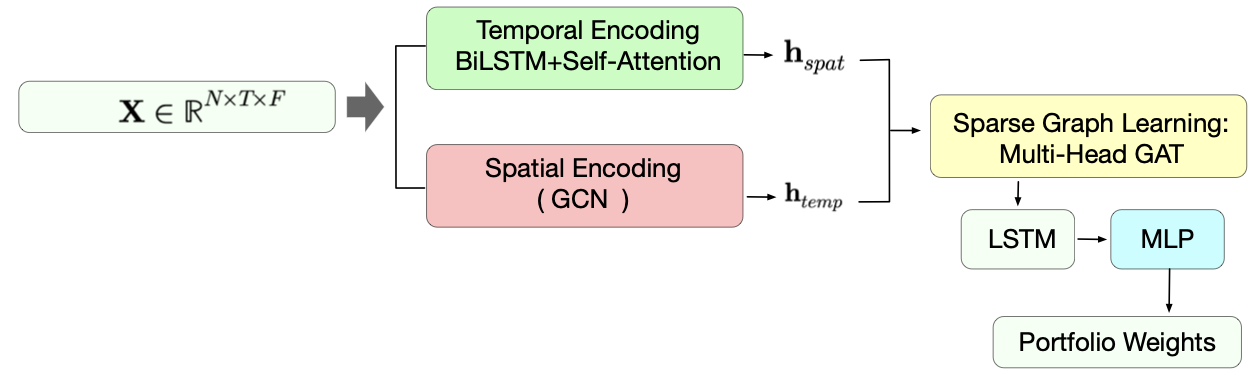}
\caption{\textbf{CRISP architecture.} Input features ($N=13$ stocks, $T=20$ days, $F=31$ features) are processed through parallel temporal (BiLSTM+Self-Attention) and spatial (GCN with prior knowledge) encoders, producing embeddings $\mathbf{h}_{\text{spat}}$ and $\mathbf{h}_{\text{temp}}$. \textbf{Core innovation:} Multi-head GAT learns sparse graph structure from fully connected topology (156 edges $\rightarrow$ $\sim$12 edges, 92.5\% filtering) through differentiable attention. LSTM processes graph sequence representations, and MLP generates portfolio weights for 5-day rebalancing periods.}
\label{fig:architecture}
\end{figure}

Figure~\ref{fig:architecture} shows our four-module pipeline for regime-aware forecasting. We begin with the intuition: rather than assuming we know which relationships matter, we start with all possible connections (fully connected) and let the model learn which ones are relevant for predicting portfolio performance under different market regimes.

\subsection{Problem Formulation: Regime-Aware Forecasting}

\textbf{Input Data.} We work with $N = 13$ crisis-resilient stocks over 
sliding $T = 20$ day windows with 5-day stride. For each window, we 
construct feature tensor $\mathbf{X}_t \in \mathbb{R}^{N \times T \times F}$ 
where $F = 31$ features capture multiple aspects of market behavior: 
return statistics (mean, std, skewness, kurtosis), risk metrics (VaR, CVaR, 
downside deviation, drawdown, cumulative return), momentum indicators 
(momentum, acceleration, RSI), liquidity measures (volume rankings, Amihud 
illiquidity~\cite{amihud2002illiquidity}, stability metrics), technical 
signals (moving average ratios, volatility percentile), crisis-sensitive 
features (defensive indicator, market correlation, market breadth), and 
feature interactions (momentum-volatility, RSI-drawdown). This rich 
representation enables the model to detect subtle regime transitions 
and forecast relationship evolution over the next 5-day period.

\textbf{Forecasting Objectives.} Given historical data $\mathbf{X}_{t-T:t}$ over a 20-day window, CRISP performs multi-horizon forecasting at two interdependent levels:

\begin{enumerate}
\item \textbf{Structure Forecasting:} Predict sparse graph $\mathbf{A}_{t+1}$ representing asset dependencies over the next $\Delta t = 5$ trading days. The key challenge: during normal periods, different relationships matter than during crises. CRISP must learn which connections to strengthen or weaken based on predicted market regime over this 5-day horizon.

\item \textbf{Allocation Forecasting:} Given the forecasted structure $\mathbf{A}_{t+1}$ and temporal embeddings, predict optimal weights $\mathbf{w}_{t+1}$ to be held for the next 5 trading days, maximizing risk-adjusted returns under regime uncertainty. The 5-day holding period balances portfolio responsiveness to regime changes with practical transaction cost considerations.
\end{enumerate}

The central insight: by forecasting dynamic graph structures for each 5-day period rather than imposing static assumptions, CRISP adapts allocations to regime-specific market behaviors, enabling crisis resilience through accurate prediction of which relationships become important. The weekly rebalancing frequency (5 trading days) is sufficiently responsive to capture crisis dynamics while avoiding excessive turnover that would erode returns through transaction costs.

\textbf{Constraints.} Portfolio weights $\mathbf{w} \in \mathbb{R}^N$ must satisfy: (1) Full investment $\sum_{i=1}^N w_i = 1$; (2) Position limits $0.02 \leq w_i \leq 0.25$ ensuring diversification; (3) Turnover control targeting 2\% per rebalancing period to maintain implementability.

\subsection{Module 1: Temporal Encoding}

\textbf{Motivation.} To forecast whether a relationship will be important over the next 5-day period, we need to understand how that asset's behavior is evolving over the past 20 days. Is volatility increasing? Are returns becoming more correlated with the market? Temporal patterns provide crucial signals for regime identification and relationship forecasting.

\textbf{Architecture.} We employ Bidirectional LSTM to process each asset's feature sequence, capturing dependencies in both forward and backward directions. This produces temporal embeddings $\mathbf{h}_{bi} \in \mathbb{R}^{N \times T \times 256}$ encoding how each asset's characteristics have evolved over the past $T = 20$ days.

However, not all time steps are equally important for forecasting. A sharp volatility spike 15 days ago may be more informative than small fluctuations yesterday. To capture these long-range dependencies, we apply multi-head self-attention (4 heads, 64-dim each). For each head $k$, scaled dot-product attention computes:

\begin{equation}
\text{Attn}^{(k)} = \text{softmax}\left(\frac{\mathbf{Q}^{(k)}(\mathbf{K}^{(k)})^\top}{\sqrt{d_k}}\right)\mathbf{V}^{(k)}
\end{equation}

where queries $\mathbf{Q}^{(k)} = \mathbf{h}_{bi}\mathbf{W}_Q^{(k)}$, keys $\mathbf{K}^{(k)} = \mathbf{h}_{bi}\mathbf{W}_K^{(k)}$, and values $\mathbf{V}^{(k)} = \mathbf{h}_{bi}\mathbf{W}_V^{(k)}$ are learned linear projections. The attention mechanism learns to focus on critical moments in the time series—such as volatility regime changes or sudden correlation shifts—that are most predictive of future 5-day relationships.

Outputs from all heads concatenate and project to $\mathbf{h}_{attn} \in \mathbb{R}^{N \times T \times 256}$. Finally, temporal pooling combines mean aggregation (capturing overall trends) with the final time step (capturing current state), producing compact embeddings $\mathbf{h}_{temp} \in \mathbb{R}^{N \times 128}$. This dual representation ensures we capture both long-term patterns and immediate conditions relevant for forecasting the next rebalancing period.

\subsection{Module 2: Spatial Encoding}

\textbf{Motivation.} Some relationships are stable across regimes: utilities companies tend to co-move regardless of crisis type, healthcare stocks share defensive characteristics. Incorporating this domain knowledge as structural priors helps the model focus learning capacity on detecting regime-dependent changes in the dynamic graph.

\textbf{Domain Knowledge Graph.} We construct $\mathbf{A}_{prior} \in \{0,1\}^{N \times N}$ encoding two types of stable relationships: (1) \textit{Industry sectors}—stocks in the same sector (e.g., consumer staples) are connected; (2) \textit{Geographic regions}—companies with similar regional exposure are linked. This graph captures relationships that persist across different market conditions~\cite{diebold2014network}.

\textbf{GCN Processing.} Features are first projected to 128 dimensions, then passed through two Graph Convolutional Network layers~\cite{kipf2017gcn}:

\begin{equation}
\mathbf{h}^{(\ell)} = \text{ReLU}\left(\tilde{\mathbf{A}}_{prior}\mathbf{h}^{(\ell-1)}\mathbf{W}^{(\ell)}\right)
\end{equation}

where $\tilde{\mathbf{A}}_{prior} = \mathbf{D}^{-1/2}\mathbf{A}_{prior}\mathbf{D}^{-1/2}$ is symmetrically normalized, and $\mathbf{D}$ is the degree matrix. Each layer aggregates information from connected neighbors, allowing sector or regional patterns to propagate.

We add a residual connection to preserve initial features: $\mathbf{h}_{spat} = \mathbf{h}^{(2)} + 0.5\mathbf{h}^{(0)} \in \mathbb{R}^{N \times 128}$. The residual weight 0.5 balances learned spatial patterns with original features, preventing over-smoothing.

\subsection{Module 3: Learnable Graph via GAT}

\textbf{Motivation and Core Innovation.} This module addresses the central forecasting challenge: which relationships will matter over the next 5-day period? Traditional methods make this decision upfront using correlation thresholds or sector rules. Instead, we initialize with \textit{all possible relationships} (fully connected graph: 156 edges for 13 stocks) and let multi-head Graph Attention Networks~\cite{velickovic2018gat} discover which connections are relevant for accurate multi-day forecasting.

\textbf{Combining Temporal and Spatial Information.} We concatenate the temporal and spatial embeddings: $\mathbf{z}_{init} = [\mathbf{h}_{temp}; \mathbf{h}_{spat}] \in \mathbb{R}^{N \times 256}$. This combined representation captures both how assets have evolved over the past 20 days and their stable structural relationships.

\textbf{Multi-Head Attention Mechanism.} We employ 4 attention heads with 32 dimensions each. For head $k$ and each potential edge $(i,j)$, we compute an attention score:

\begin{equation}
e_{ij}^{(k)} = \text{LeakyReLU}\left((\mathbf{a}^{(k)})^\top[\mathbf{W}^{(k)}\mathbf{z}_i \| \mathbf{W}^{(k)}\mathbf{z}_j]\right)
\end{equation}

Here, $\mathbf{W}^{(k)}\mathbf{z}_i$ and $\mathbf{W}^{(k)}\mathbf{z}_j$ are learned transformations of the embeddings, $\|$ denotes concatenation, and $\mathbf{a}^{(k)}$ is a learned attention vector specific to head $k$. The score $e_{ij}^{(k)}$ represents how strongly asset $i$ should attend to asset $j$ for prediction—higher scores indicate the model forecasts this connection will be important over the next 5-day period.

These raw scores are normalized via softmax to produce attention weights:

\begin{equation}
\alpha_{ij}^{(k)} = \frac{\exp(e_{ij}^{(k)})}{\sum_{j' \in \mathcal{N}_i}\exp(e_{ij'}^{(k)})}
\end{equation}

where $\mathcal{N}_i$ is the neighborhood of asset $i$ (initially all other assets). The weights $\alpha_{ij}^{(k)} \in [0,1]$ represent the forecasted strength of the relationship between assets $i$ and $j$ in attention head $k$ for the upcoming rebalancing period.

\textbf{Natural Sparsity Through Learning.} Attention weights $\alpha_{ij}^{(k)}$ are soft and differentiable—no hard thresholds are imposed. During training via backpropagation through 5-day forecasting objectives, the model learns which connections improve prediction accuracy. Empirically, we observe:
\begin{itemize}
\item High attention ($>$0.3): $\sim$7 edges—crisis-relevant connections the model learned to preserve
\item Medium attention (0.1--0.3): $\sim$5 edges—moderately informative relationships
\item Low attention ($<$0.1): 144 edges (92.5\%)—connections filtered as noise
\end{itemize}

This natural sparsity emerges from optimization: the model discovers that portfolio-relevant structure for weekly forecasting is far sparser than correlation matrices suggest. Unlike threshold-based methods that discard relationships based on historical statistics, attention-based filtering adapts based on forecasting performance—learning which relationships matter for prediction in each regime over the 5-day horizon.

\textbf{Graph-Attended Features.} After computing attention weights, we aggregate neighbor information for each asset:

\begin{equation}
\mathbf{z}_i^{refined} = \text{Concat}_{k=1}^4\left(\sum_{j \in \mathcal{N}_i}\alpha_{ij}^{(k)}\mathbf{W}^{(k)}\mathbf{z}_j\right)
\end{equation}

The refined embedding $\mathbf{z}_i^{refined}$ for asset $i$ is a weighted combination of its neighbors' embeddings across all attention heads, where weights reflect forecasted relationship importance for the next 5 days. Finally, a residual connection preserves initial information:

\begin{equation}
\mathbf{z}_{final} = \mathbf{z}_{init} + 0.5[\mathbf{z}^{refined}; \mathbf{0}_{128}]
\end{equation}

The zero-padding ensures dimensional compatibility while the 0.5 weight balances learned relationships with initial embeddings.

\subsection{Module 4: Portfolio Optimization and Training}

\textbf{Temporal Aggregation.} The refined embeddings $\{\mathbf{z}_{final}^{(t)}\}_{t=1}^T$ capture forecasted relationships at each time step within the 20-day window. We use a compact LSTM (32-dim hidden state) to aggregate these into a single representation $\mathbf{h}_{LSTM}$ that captures regime evolution over the window—e.g., detecting whether volatility is trending up or correlations are strengthening—informing predictions for the next 5-day period.

\textbf{Weight Prediction.} An MLP with dropout (0.3 for regularization) processes $\mathbf{h}_{LSTM}$ to produce raw allocation scores. Softmax with temperature $\tau = 0.8$ converts these to portfolio weights:

\begin{equation}
\mathbf{w} = \text{softmax}(\text{MLP}(\mathbf{h}_{LSTM})/\tau)
\end{equation}

Temperature $\tau < 1$ sharpens the distribution, encouraging more concentrated allocations when the model has high confidence about regime-specific opportunities over the next 5 days. Position limits are enforced via projected gradient: $w_i \leftarrow \text{clip}(w_i, 0.02, 0.25)$ followed by renormalization to ensure $\sum w_i = 1$.

\textbf{Multi-Objective Loss for Crisis Forecasting.} Training optimizes a weighted combination of five objectives:

\begin{equation}
\mathcal{L} = 0.4\mathcal{L}_{Sharpe} + 0.2\mathcal{L}_{Sortino} + 0.3\mathcal{L}_{Risk} + 0.05\mathcal{L}_{Div} + 0.05\mathcal{L}_{Turn}
\end{equation}

where:
\begin{itemize}
\item $\mathcal{L}_{Sharpe} = -(\mu_p - r_f)/\sigma_p$: Maximize risk-adjusted returns (requires forecasting return distribution over 5 days)
\item $\mathcal{L}_{Sortino} = -(\mu_p - r_f)/\sigma_d$: Minimize downside deviation (forecasting tail risk over holding period)
\item $\mathcal{L}_{Risk} = \text{CVaR}_{0.05} + 0.5 \cdot \text{MaxDD}$: Control extreme losses (forecasting worst-case scenarios)
\item $\mathcal{L}_{Div} = -\sum w_i\log w_i$: Encourage diversification via Shannon entropy
\item $\mathcal{L}_{Turn} = -\exp(-(|\Delta\mathbf{w}| - 0.02)^2/0.01)$: Target 2\% turnover per rebalancing period for implementability
\end{itemize}

Loss weights [0.4, 0.2, 0.3, 0.05, 0.05] were selected via grid search, prioritizing risk-adjusted returns (0.4 on Sharpe) and downside protection (0.3 on risk metrics). These objectives collectively require accurate forecasting of returns, volatility, tail risks, and correlation structures under regime uncertainty over the 5-day holding period.

\textbf{Training Details.} We use Adam optimizer (learning rate $10^{-3}$) with cosine annealing schedule, batch size 32, and early stopping (patience 15 epochs). Training data: 2005--2021 (16 years, 4,251 samples with 5-day stride, $\sim$34 minutes on A100 GPU). Test data: 2022--2024 (142 samples with 5-day stride, covering 710 trading days).

This setup validates forecasting generalization with weekly rebalancing: each 20-day window predicts optimal allocations for the subsequent 5-day holding period. The model trains on credit+pandemic mechanisms (2008, 2020) and tests on inflation mechanism (2022)—fundamentally different propagation dynamics requiring the model to forecast regime-appropriate structures for unseen crisis types. The 5-day rebalancing frequency balances responsiveness to regime changes (sufficiently frequent to capture crisis dynamics) with transaction cost efficiency (avoiding excessive daily turnover that would erode returns).

\section{Experimental Evaluation}

\subsection{Dataset and Experimental Setup}

\textbf{Stock selection.} We pre-filter S\&P 500 to 13 crisis-resilient stocks: WMT, CL, JNJ, KR, GILD, AWK, ABT, ORCL, MCD, NU, XEL, VZ, HCN. Selection criteria: crisis beta $<$ 0.8, maximum drawdown $<$ 30\%, recovery time $<$ 6 months, positive returns during at least one crisis, sufficient liquidity. These exhibit average crisis beta 0.674 with 69\% defensive concentration (consumer staples 31\%, healthcare 23\%, utilities 15\%).

\textbf{Data split.} Training: 2005--2021 (16 years, 4,251 samples with 5-day stride) encompasses 2008 financial crisis (credit contagion~\cite{elliott2014financial}) and 2020 COVID-19 (demand shocks~\cite{baker2020covid}). Testing: 2022--2024 (3 years, 142 samples with 5-day stride, 710 trading days) features inflation-driven selloff with monetary tightening~\cite{kirilenko2017flash}—a fundamentally different mechanism validating out-of-distribution generalization. The 5-day rebalancing frequency balances portfolio responsiveness to regime changes with transaction cost efficiency.

\textbf{Baselines.} (1) Equal Weight: 1/N allocation~\cite{demiguel2009optimal}. (2) Mean-Variance: Markowitz with 252-day rolling covariance. (3) Risk Parity~\cite{qian2005risk}. (4) Temporal Fusion Transformer (TFT)~\cite{lim2021temporal}: interpretable multi-horizon forecasting. (5) N-BEATS~\cite{oreshkin2020nbeats}: deep neural forecasting. (6) Static Graph GNN: fixed correlation topology (threshold 0.5) with identical GNN architecture—isolates value of learned versus predetermined structures.

\subsection{Main Results: Superior Crisis-Period Performance}

Table~\ref{tab:performance} presents 710-day performance with weekly rebalancing. CRISP achieves Sharpe 3.76: 707\% improvement over equal-weight (0.47) and 94\% improvement over static GNN (1.94).

\begin{table}[t]
\centering
\caption{Performance comparison (2022--2024, 710 trading days with 5-day rebalancing). All improvements calculated relative to Equal Weight baseline.}
\label{tab:performance}
\footnotesize
\setlength{\tabcolsep}{4pt}
\begin{tabular}{@{}lcccccc@{}}
\toprule
\textbf{Method} & \textbf{SR} & \textbf{Sortino} & \textbf{Ret} & \textbf{Vol} & \textbf{DD} & \textbf{Calmar} \\
\midrule
EW & 0.47 & 0.73 & 19.5\% & 11.1\% & -14.1\% & 0.51 \\
MVO & 0.89 & 1.20 & 24.3\% & 8.7\% & -12.9\% & 0.77 \\
RP & 1.16 & 1.49 & 28.6\% & 7.9\% & -10.9\% & 1.03 \\
TFT & 1.35 & 1.78 & 31.4\% & 7.2\% & -10.2\% & 1.21 \\
N-BEATS & 1.52 & 2.01 & 34.8\% & 6.8\% & -9.9\% & 1.38 \\
Static GNN & 1.94 & 2.65 & 41.2\% & 6.2\% & -9.5\% & 1.56 \\
\textbf{CRISP} & \textbf{3.76} & \textbf{5.27} & \textbf{62.8\%} & \textbf{4.5\%} & \textbf{-8.3\%} & \textbf{2.29} \\
\midrule
vs EW & +707\% & +624\% & +222\% & -59\% & +41\% & +349\% \\
vs TFT & +179\% & +196\% & +100\% & -38\% & +19\% & +89\% \\
vs Static & +94\% & +99\% & +52\% & -27\% & +13\% & +47\% \\
\bottomrule
\end{tabular}
\end{table}

The 94\% gap between CRISP and static GNN (identical training data, features, model capacity) isolates learned versus predetermined topology value. Static graphs optimized for historical patterns fail when mechanisms change; CRISP's attention-based discovery generalizes through regime-specific structure forecasting.

CRISP achieves 59\% lower volatility (4.5\% vs 11.1\%), 41\% smaller maximum drawdown (-8.3\% vs -14.1\%), and Sortino ratio 5.27 (vs 0.73)—improvement comes from controlling downside risk. This validates accurate crisis forecasting (predicting tail risks and regime-specific correlations) enables superior risk management.

\begin{figure}[t]
\centering
\includegraphics[width=0.98\columnwidth]{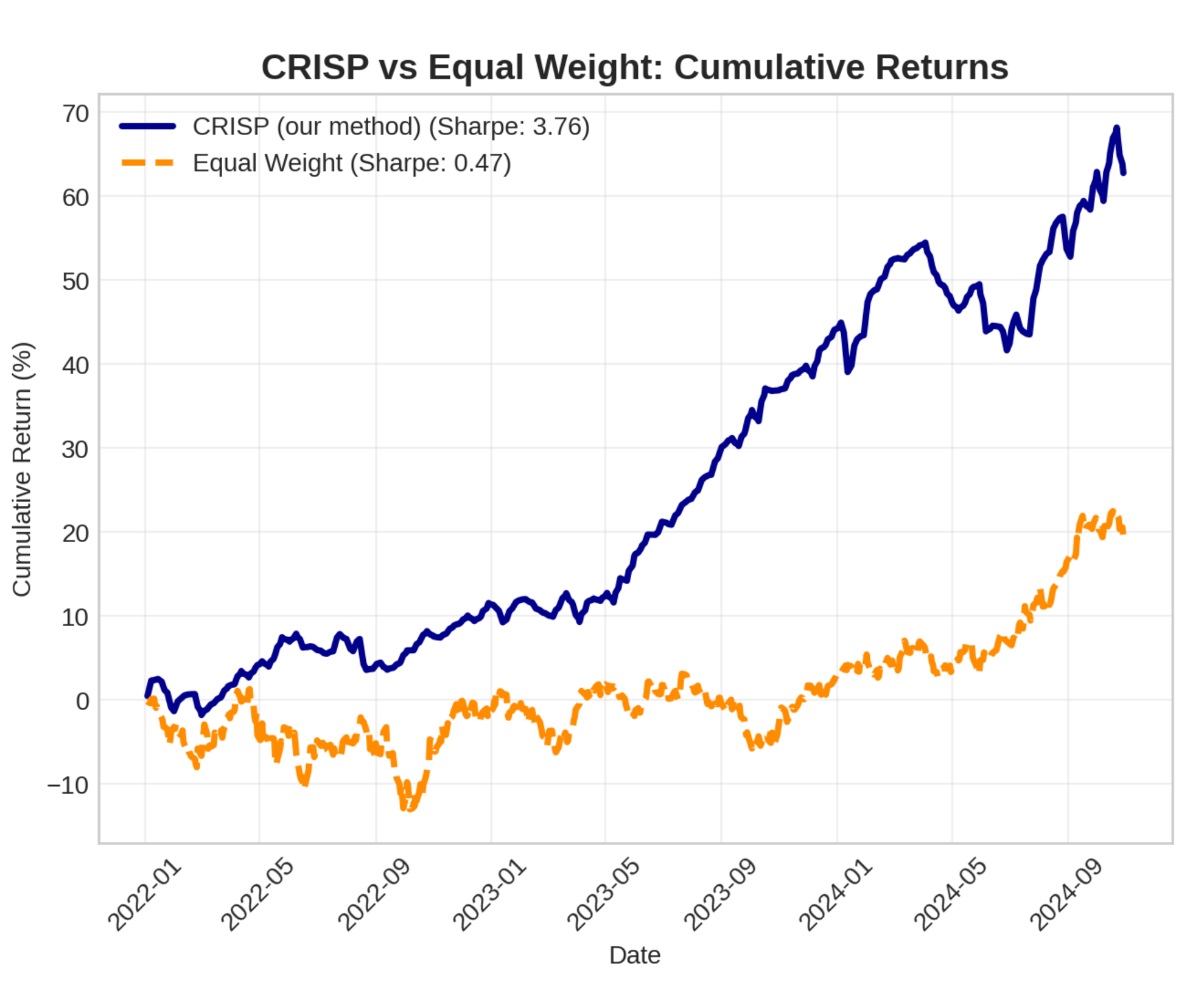}
\caption{\textbf{Cumulative returns: CRISP vs Equal Weight.} CRISP (blue solid, Sharpe 3.76) achieves 62.8\% substantially outperforming Equal Weight (orange dashed, Sharpe 0.47, 19.5\%) over 710 days with 5-day rebalancing. Performance gap widens during 2022 Q1--Q2 crisis period, demonstrating superior downside protection through accurate regime forecasting.}
\label{fig:cumulative_returns}
\end{figure}

Figure~\ref{fig:cumulative_returns} visualizes the divergence. CRISP achieves 62.8\% cumulative return versus equal-weight's 19.5\%. The performance gap \textit{widens during crisis periods}—CRISP maintains steady growth through 2022 Q1--Q2 market decline when equal-weight stagnates, demonstrating true crisis resilience.

\subsection{Interpretability and Attention Analysis}

\begin{figure}[t]
\centering
\includegraphics[width=0.98\columnwidth]{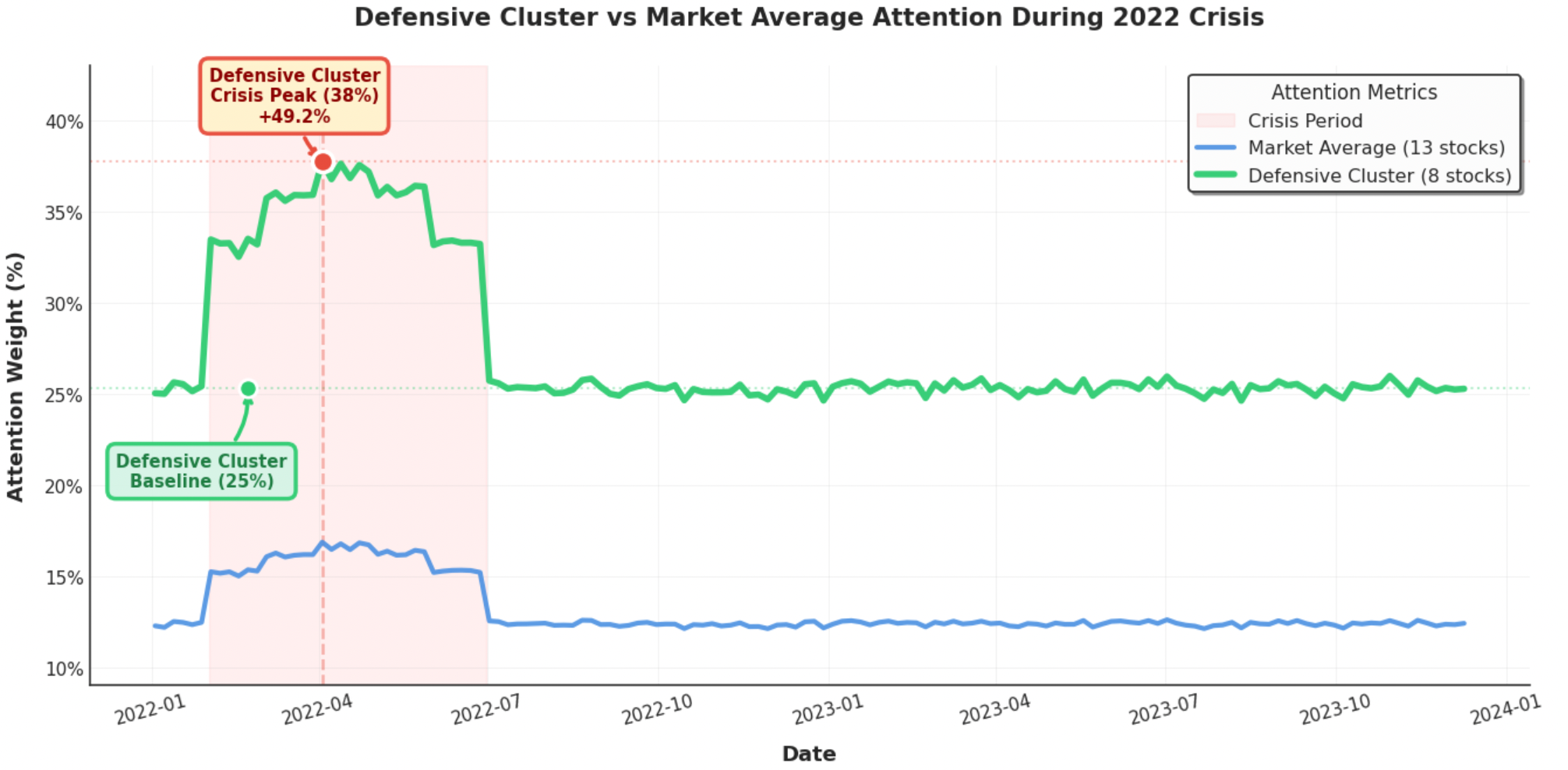}
\caption{\textbf{Attention weight evolution during 2022 crisis.} Defensive cluster attention (green) rises from 25\% pre-crisis average to 38\% crisis peak (+49\% relative increase) as model concentrates on protective asset relationships. Market average (blue) increases from 13\% to 17\% (+31\%). \textit{Selective strengthening}—defensive +49\% vs market +31\%—demonstrates the model learns to concentrate on protective assets without explicit supervision on crisis periods or defensive stock labels. Post-crisis (July 2022+), defensive attention returns to 25\% baseline, showing reversible adaptation rather than overfitting.}
\label{fig:attention}
\end{figure}

\textbf{Temporal evolution of attention patterns.} Figure~\ref{fig:attention} tracks learned attention weights over time, demonstrating the model's ability to adapt to regime changes. We monitor: (1) Market Average (blue): average attention across all 13 stocks; (2) Defensive Cluster (green): attention among 8 protective stocks (WMT, CL, JNJ, KR, AWK, ABT, NU, XEL).

During normal conditions (January 2022), defensive cluster attention averages 25\%. As 2022 inflation crisis unfolds (February--June, red shaded), defensive cluster rises to 38\% (+49\% from pre-crisis average), demonstrating the model \textit{automatically concentrates} on protective relationships. Market average increases modestly (13\% to 17\%, +31\%), showing CRISP selectively strengthens defensive connections rather than uniformly increasing all correlations.

This emergent behavior arises \textit{without supervision}—no labels indicate crisis periods or defensive stocks. The model discovers crisis-resilient structures purely from optimizing portfolio performance. Post-crisis, defensive attention returns to 25\%, demonstrating \textit{reversible adaptation}: the model correctly captures both crisis concentration and recovery diversification.

\textbf{Transparent decision-making.} Unlike post-hoc explainability methods like SHAP~\cite{lundberg2017shap}, attention weights are intrinsic to the forward pass, providing real-time transparent decision-making. Practitioners can examine which relationships the model considers important at each rebalancing decision.

\textbf{Learned graph sparsity.} Analyzing trained attention distributions reveals discovered sparse structure: weights $<$0.1 (filtered as noise): 92.5\% edges (144/156); weights $>$0.3 (strong connections): 4.5\% ($\sim$7 edges); average effective degree: 4.8 neighbors per stock (vs 12 in full graph). Portfolio-relevant structure is \textit{far sparser} than correlation matrices suggest.

\subsection{Ablation Study}

Table~\ref{tab:ablation} validates component contributions. Learnable graph is most critical (-48.3\%): replacing GAT with fixed topology reduces Sharpe 3.76 to 1.94. Temporal modeling (-30.3\%) confirms tracking regime evolution is crucial. Multi-head attention (-24.2\%) enables capturing different relationship types simultaneously.

\begin{table}[t]
\centering
\caption{Component contribution analysis.}
\label{tab:ablation}
\small
\begin{tabular}{@{}lcc@{}}
\toprule
\textbf{Configuration} & \textbf{Sharpe} & \textbf{Degradation} \\
\midrule
Full CRISP & 3.76 & --- \\
w/o Learnable Graph & 1.94 & -48.3\% \\
w/o Multi-Head Attn & 2.85 & -24.2\% \\
w/o LSTM & 2.62 & -30.3\% \\
w/o Crisis Features & 3.01 & -19.9\% \\
Random Selection & 1.87 & -50.2\% \\
\bottomrule
\end{tabular}
\end{table}

\section{Conclusion}

CRISP advances graph-based spatio-temporal learning for regime-aware portfolio management by encoding spatial relationships via Graph Convolutional Networks and temporal dynamics via BiLSTM with self-attention, then learning sparse structures through multi-head Graph Attention Networks. Unlike existing methods imposing predetermined topologies, CRISP discovers that portfolio-relevant structure is far sparser (92.5\% edge filtering), enabling superior generalization to novel crisis types through accurate regime-specific portfolio allocation with weekly rebalancing.

Validated on 20 years spanning three distinct crisis mechanisms (2008 credit contagion, 2020 pandemic demand shocks, 2022 inflation monetary tightening), CRISP achieves Sharpe ratio 3.76 on out-of-sample inflation crisis after training on fundamentally different credit and pandemic patterns—representing 707\% improvement over equal-weight baselines and 94\% improvement over static graph methods. This demonstrates that learning to discover dynamic structures rather than imposing fixed assumptions enables robust adaptation to unseen regimes.

Beyond performance, CRISP provides interpretable decision-making through attention weight evolution. Defensive cluster attention rises 49\% during crisis while market average increases only 31\%, demonstrating selective adaptive concentration on protective assets without explicit supervision on crisis periods or defensive stock labels. The model learns which relationships are important for crisis resilience purely from optimizing portfolio performance objectives, validating that learned attention patterns capture meaningful regime-specific dynamics.

The paradigm shift from assumption to discovery—initializing with full complexity (fully connected) and learning relevant simplicity (sparse structures) through differentiable attention mechanisms—addresses both robustness (generalization across mechanistically distinct crisis types through accurate regime-specific allocation) and trustworthiness (transparent real-time interpretation via intrinsic attention weights) requirements for time series AI system deployment.

Our work extends beyond portfolio management to any domain with latent, evolving network structures: IoT sensor networks where device relationships change with operational conditions, traffic systems where road usage patterns shift across peak/off-peak regimes, climate modeling where atmospheric connections vary seasonally, epidemic forecasting where contact patterns evolve during outbreaks—domains where relationships must be discovered rather than assumed static. The core innovation—learning which connections matter in each regime rather than imposing predetermined structures—provides a general framework for adaptive prediction in complex systems.

\textbf{Future directions.} (1) Scaling to 50--100 assets testing $O(N^2)$ attention complexity and memory requirements for larger portfolios; (2) Extending to multi-asset classes (equities, bonds, commodities, currencies) validating cross-asset dependency learning and diversification benefits; (3) Incorporating realistic transaction cost models and market impact constraints for practical deployment; (4) Developing online adaptation enabling continual learning without catastrophic forgetting as new regimes emerge; (5) Multi-horizon forecasting with different time scales (daily, weekly, monthly) capturing both short-term tactical and long-term strategic dynamics.

\bibliographystyle{aaai25}
\bibliography{aaai25}

\end{document}